%% file: improve360.tex
\documentclass[10pt,twocolumn,letterpaper]{article}

\usepackage{cvpr}
\usepackage{times}
\usepackage{epsfig}
\usepackage{graphicx}
\usepackage{amsmath}
\usepackage{amssymb}

\usepackage{multirow}
\usepackage{gensymb}
\usepackage{cite}
\usepackage{enumitem}
\usepackage{array}
\usepackage{subcaption}
\usepackage[font=small]{caption}
\usepackage{textcomp}

\usepackage[breaklinks=true,bookmarks=false]{hyperref}

\cvprfinalcopy 

\def\cvprPaperID{2894} 

\ifcvprfinal\fi
\begin{document}

\title{Making 360\textdegree~Video Watchable in 2D:\\Learning Videography for Click Free Viewing}

\author{
Yu-Chuan Su
\hspace{4em}
Kristen Grauman\\
The University of Texas at Austin
}

\maketitle

\begin{abstract}
360\textdegree~video requires human viewers to actively control ``where'' to look while watching the video.
Although it provides a more immersive experience of the visual content, it also introduces additional burden for viewers;
awkward interfaces to navigate the video lead to suboptimal viewing experiences.
Virtual cinematography is an appealing direction to remedy these problems,
but conventional methods are limited to virtual environments or rely on hand-crafted heuristics.
We propose a new algorithm for virtual cinematography that automatically controls a virtual camera within a 360\textdegree~video.
Compared to the state of the art, our algorithm allows more general camera control,
avoids redundant outputs, and extracts its output videos substantially more efficiently.
Experimental results on over 7 hours of real ``in the wild'' video show that our generalized camera control is crucial for viewing 360\textdegree~video,
while the proposed efficient algorithm is essential for making the generalized control computationally tractable.
\vspace{-0.1in}
\end{abstract}

\section{Introduction}
\vspace{-0.02in}

\input{introduction}

\vspace{-0.04in}

\section{Related Work}

\input{related}

\vspace{-0.02in}

\section{Preliminaries: \textsc{AutoCam}}
\label{sec:preliminaries_autocam}
\vspace{-0.04in}

\input{autocam}

\vspace{-0.04in}
\section{Approach}
\vspace{-0.04in}

In this section,
we present our algorithm.
First we generalize the original Pano2Vid problem to enable zooming (Sec.~\ref{sub:camera_zoom_in_pano2vid}).
Next, we describe our coarse-to-fine approach to reduce the computational cost of camera trajectory selections (Sec.~\ref{sub:two_stage_trajectory_search}).
Finally, we introduce an iterative approach to generate a diverse set of output trajectories (Sec.~\ref{sub:diverse_trajectory_search}).

\subsection{Zoom Lens Pano2Vid}
\label{sub:camera_zoom_in_pano2vid}
\vspace{-0.04in}

\input{zoom_lens}

\subsection{Coarse-to-Fine Camera Trajectory Search}
\label{sub:two_stage_trajectory_search}
\vspace{-0.04in}

\input{two_stages}

\subsection{Diverse Camera Trajectory Search}
\label{sub:diverse_trajectory_search}

\input{diverse_search}

\vspace{-0.04in}
\section{Experiments}
\vspace{-0.02in}

Next we validate our method on challenging videos.
See project webpage for video examples and comparisons\footnote{\label{note:webpage}\url{http://vision.cs.utexas.edu/projects/watchable360/}}.

\vspace{-0.04in}
\subsection{Dataset}
\label{sub:dataset}
\vspace{-0.04in}

\input{dataset}

\vspace{-0.02in}
\subsection{Baselines}
\label{sub:baselines}
\vspace{-0.04in}

\input{baselines}

\vspace{-0.12in}
\subsection{Evaluation Metrics}
\label{sub:evaluation_metrics}
\vspace{-0.04in}

\input{metrics}

\vspace*{-0.04in}
\subsection{Output Quality}
\label{sub:output_quality}
\vspace*{-0.02in}

\input{output_quality}

\vspace*{-0.02in}
\subsection{Computational Cost}
\label{sub:computational_cost}
\vspace*{-0.02in}

\input{computational_cost}

\vspace*{-0.04in}
\subsection{Output Diversity}
\label{sub:output_diversity}
\vspace*{-0.02in}

\input{output_diversity}

\vspace*{-0.04in}
\section{Conclusion}
\vspace*{-0.04in}

\input{conclusion}

\vspace{0.08in}
\noindent {\bf Acknowledgement}.
We thank Chia-Chen, Kimberly, Ruohan, Wei-Ju, and YiHsuan for collecting annotation.
This research is supported in part by NSF IIS -1514118 and a gift from Intel.

{\small
\bibliographystyle{ieee}
\bibliography{improve360}
}

\clearpage

\appendix

\input{supp}

\end{document}

%% file: introduction.tex
$360\degree$ cameras are becoming very popular,
thanks to emerging Virtual Reality (VR) technologies and applications.
More than a dozen new $360\degree$ cameras were released in 2016~\cite{new360cam},
and the market is expected to grow by more than $100\%$ per year in the next few years~\cite{cta2016research}.
Watching panoramic photos and videos is becoming a common experience on content distribution sites like YouTube and Facebook,
and many content creators are adopting the new medium.
For example, BBC News distributes news in $360\degree$ videos online~\cite{bbc360news}.
Whereas traditional cameras are restricted to a field of view (FOV) even narrower than human perception,
a $360\degree$ camera captures the entire visual world from its optical center.

\begin{figure}[t]
    \center
    \includegraphics[width=1.\linewidth]{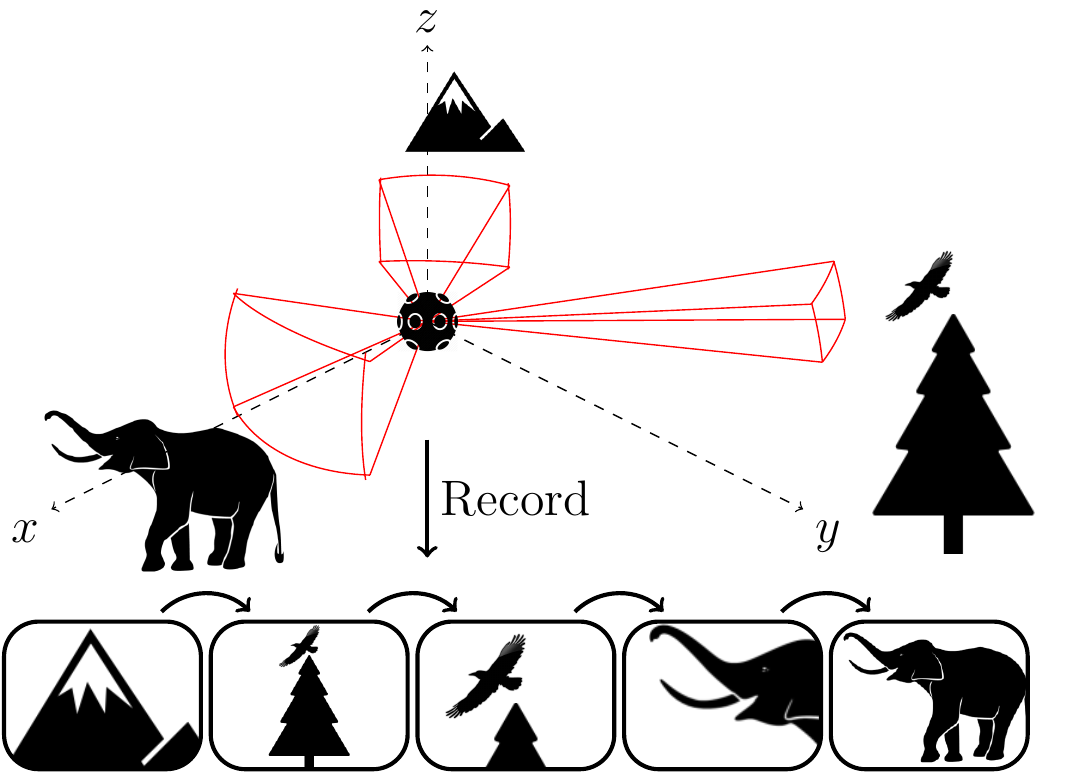}
    \vspace{-15pt}
    \caption{
        Our goal is to control the direction and field of view of a virtual camera within a $360\degree$ video in order to record a video that looks as if it were captured by a human videographer.
    }
    \label{fig:concept}
    \vspace{-12pt}
\end{figure}

This broad capture offers new freedom to videographers and video consumers alike.
A videographer no longer has to determine which direction to capture in the scene,
freeing her to experience the moment rather than the act of recording a video.
Meanwhile, a human video consumer has the freedom to explore the visual content based on her interest, without being severely restricted by choices made by the videographer.
For example, a news correspondent can traverse a war zone without consciously considering how to portray the scene,
and subsequent viewers will still have an immersive experience about the tragedy and witness events in more detail than the videographer may even be able to attend to in the moment.
Similarly, a parent at his child's bustling birthday party can passively record $360\degree$ memories to be perused more deliberately later.

On the other hand, the medium also introduces new challenges.
The most common interface for watching $360\degree$ videos is to display a small portion of the video as a 2D normal field of view (NFOV) video captured by a virtual camera\footnote{For example, try clicking around on \url{https://www.youtube.com/watch?v=aTTzKwLPqFw}}.
The video viewer now has to decide ``where and what'' to look at by controlling the direction of the virtual camera throughout the full duration of the video.
The display can be a normal screen, a mobile device, or a VR headset,
and the control signal will come from the mouse, the pose of the device, or the head movement respectively.
These choices determine the content seen by the viewer and thus the user experience.
Because the viewer has no information about the content beyond the current FOV,
it may be difficult to determine where to look in real time,
e.g., a $360\degree$ video viewer can easily fail to notice that there is something approaching the camera from the opposite direction.
In fact, the viewer may have to watch the video multiple times in order to find a proper way to control the virtual camera that navigates through the content of interest.
While $360\degree$ videos may alternatively be displayed in their entirety using equirectangular projection, the unfamiliar format and distortion make such video hard to watch.

In light of this challenge, $360\degree$ video is a particularly appealing domain to invoke \emph{automatic videography} techniques, which aim to convert unedited materials into
an effective video presentation that conveys events~\cite{christianson1996declarative,he1996virtual,elson2007lightweight,mindek2015automatized,foote2000flycam,sun2005region,su2016accv,heck2007videography,rui2004automating}.
While automatic videography in prior work has largely dealt with virtual environments and hand-crafted heuristics~\cite{christianson1996declarative,he1996virtual,elson2007lightweight,mindek2015automatized},
our recent work shows the potential of \emph{learning} how to extract informative portions of $360\degree$ video as a presentable NFOV video~\cite{su2016accv}.
In~\cite{su2016accv},
we introduce the \emph{Pano2Vid} problem, which takes a $360\degree$ video as input and as output generates NFOV videos that look like they could have been captured by a human observer equipped with a real NFOV camera.
Our \textsc{AutoCam} algorithm learns videography tendencies directly from human-captured Web videos.
By controling the pose of a virtual NFOV camera within the input video,
it removes the burden of deciding ``where'' to look when watching $360\degree$ videos.

We propose a new algorithm for the Pano2Vid problem that broadens the scope of camera control to generate more realistic videos.
First, we generalize the task of Pano2Vid to allow not only spatial selections within the $360\degree$ video but also changes in the FOV.
This allows the virtual camera control to fully mimic human videographer tendencies,
because changing the FOV, i.e.,~zooming, is a common technique in both professional and amateur videography.
Second, toward a more computationally efficient algorithm,
we present a coarse-to-fine search approach that iteratively refines the camera control while reducing the search space in each iteration.
The new approach makes the generalized task encompassing zoom computationally tractable.
Finally, to account for the fact that valid Pano2Vid solutions are often multimodal,
we explore how to generate a \emph{diverse} set of plausible output NFOV videos,
overcoming the redundancy that deters a straightforward optimization approach.

We experiment with more than 7 hours of real-world $360\degree$ video and
capture 12 hours of manually edited $360\degree$ data in order to
characterize results both quantitatively and qualitatively. 
We demonstrate that the proposed generalization of the Pano2Vid problem has dramatic effects.
Compared to both the existing solution as well as strong baselines driven by saliency or center biases,
our method's auto-zoom improves results by up to $43.4\%$.
In addition, we achieve a significant advantage in computational cost, reducing run-time by more than $84\%$.

%% file: related.tex
\vspace*{-0.05in}

\paragraph{Video saliency}

Saliency studies visual content that attracts viewers' attention~\cite{rudoy2013cvpr,itti-motion2003,liu2011learning,harel2006graph,itti-2009,zhai2006visual},
where attention is usually measured by gaze fixations under free viewing settings.
Although the research originated in static images,
there is increasing work that studies video saliency~\cite{itti-motion2003,zhai2006visual,rudoy2013cvpr}.
Both video saliency and Pano2Vid try to predict spatial locations in videos.
However, saliency targets locations that are eye-catching in 2D image coordinates whereas Pano2Vid predicts directions in spherical coordinates that videographers would try to capture with cameras.
Also, saliency usually depends on local image content whereas Pano2Vid depends on the content and composition of the entire FOV.

\vspace*{-0.16in}
\paragraph{Video retargeting}

Video retargeting adapts a source video by cropping and scaling to better fit the target display
while minimizing the information loss~\cite{liu2006video,avidan2007seam,rubinstein2008improved,krahenbuhl2009system,khoenkaw2015automatic,shamir-gaze-siggraph2015}.
Both retargeting and our algorithm select portions of the original video to display to the user,
but retargeting takes an already well-edited video as input and tries to generate a new version that conveys the same information.
In contrast, our input $360\degree$ video is not pre-edited,
and our goal is to generate multiple outputs that convey different information.  Pano2Vid also entails a more severe reduction in spatial extent,
e.g., compared to retargeting a 2D video from the Web to display nicely on a mobile device.

\vspace*{-0.15in}
\paragraph{Virtual cinematography}

Most existing work on virtual cinematography studies virtual camera control in virtual (computer graphics) environments~\cite{christianson1996declarative,he1996virtual,elson2007lightweight,mindek2015automatized}
or else a specialized domain such as lecture videos~\cite{foote2000flycam,sun2005region,rui2004automating}.
Aside from camera control,
some prior works also study automatic editing of raw materials like videos or photos~\cite{gleicher2000videography,gleicher2002framework,heck2007videography,social-cameras-2014}.
The goal is to generate an effective video presentation automatically to reduce human labor filming or editing.
Existing approaches typically rely on heuristics that encode popular cinematographic rules.
Our problem differs from the above in that we take unrestricted real $360\degree$ videos as input.
Furthermore, we learn the cinematography tendencies directly from Web videos.
Our work is most related to the \textsc{AutoCam} method~\cite{su2016accv},
which also treats $360\degree$ video and takes a data-driven approach.
In contrast to~\cite{su2016accv},
we generalize the problem by allowing more degrees of freedom in the camera control (i.e.~zoom) and considering diverse hypotheses.
In addition, we propose a more efficient algorithm that makes the generalized problem computationally tractable.

\vspace*{-0.15in}
\paragraph{Video summarization}
Video summarization aims to generate a concise representation for a video by removing temporal redundancy while preserving the important events~\cite{peleg,yongjae2012discovering,dale-multivideo,khosla2013large,xiong2014detecting,gong2014diverse,gygli2014creating,gygli2015video,potapov2014category,sun2014ranking}.
The goal is different from ours in the sense that video summarization selects content temporally whereas we select content spatially.
Also, the outputs of our algorithm are continuous videos that look as if they were captured by a hand-held camera in the scene,
whereas the output of a video summarization algorithm is usually keyframes or concatenated disjoint video clips.

\vspace*{-0.15in}
\paragraph{Diverse solution search}

Generating a diverse set of candidate solutions has been widely discussed in different domains~\cite{batra2012diverse,kirillov2015inferring,gimpel2013diverse,vijayakumar2016diverse,li2016mutual}.
A common approach is to find the solutions iteratively and encourage the solution of the current iteration to be different from that of previous iterations by penalizing the similarity between them.
This approach has been applied to segmentation and pose estimation in a Markov Random Field~\cite{batra2012diverse}, as well as machine translation in diverse beam search~\cite{gimpel2013diverse}.
Our approach differs in that we formulate an exact solution search using dynamic programming instead of a probabilistic model or an approximate search.
Also, our approach guarantees the minimum distance between solutions obtained in different iterations, whereas existing approaches rely on a predefined penalty term which fails to provide the same guarantee.

%% file: autocam.tex
\setlength{\abovedisplayskip}{6pt}
\setlength{\belowdisplayskip}{6pt}

First we provide background on our existing \textsc{AutoCam} solution~\cite{su2016accv}.
\textsc{AutoCam} starts by sampling spatio-temporal glimpses (ST-glimpses) from the $360\degree$ video.
An ST-glimpse is a five-second NFOV video clip with a $65.5\degree$ FOV\footnote{Note FOV refers to the horizontal field of view throughout the paper.}
 recorded from the $360\degree$ video by directing the camera to a fixed direction in the $360\degree$ camera axes.
The algorithm samples candidate ST-glimpses at 18 azimuthal angles and 11 polar angles every five seconds:
\begin{equation}
    \label{eq:sample_glimpses}
    \begin{aligned}
        \theta &\in& \Theta &= \{0,\pm10,\pm20,\pm30,\pm45,\pm75\},\\
        \phi   &\in& \Phi   &= \{0,20,\ldots,340\},\\
        t      &\in& T      &= \{0s,5s,\ldots,L-5s\},\\
    \end{aligned}
\end{equation}
where $L$ is the video length.
Each candidate ST-glimpse is defined by the camera principal axis $(\theta,\phi)$ direction and time in the video $t$:
\begin{equation}
    \Omega_{t, \theta, \phi} \equiv (\theta_t,\phi_t) \in \Theta \times \Phi.
\end{equation}
\textsc{AutoCam} then learns the ``capture-worthiness'' scores---the likelihood of a ST-glimpse appearing in human captured NFOV videos.
Based on the assumption that 1) the content in human captured NFOV videos are mostly capture-worthy and 2) most ST-glimpses are not capture-worthy,
it learns a classifier that predicts whether a video clip is a NFOV video or a ST-glimpse.
The posteriors on test ST-glimpses are their capture-worthiness scores.
The NFOV videos are crawled from YouTube and  convolutional 3D features (C3D)~\cite{tran2015learning} are used as the video representation.

After obtaining the capture-worthiness score of each candidate ST-glimpse,
\textsc{AutoCam} constructs a camera trajectory (i.e.~camera direction over time) by finding a path over the ST-glimpses that maximizes the \emph{aggregate} capture-worthiness score,
while simultaneously producing human-like smooth camera motions.
It realizes the smooth camera motion by restricting the trajectory from choosing an ST-glimpse that is displaced from the previous one by more than a threshold $\epsilon$, i.e.
\begin{equation}
    \label{eq:autocam_constraints}
    |\Delta \Omega|_{\theta} = |\theta_{t}-\theta_{t-1}| \le \epsilon_{\theta}, \;
    |\Delta \Omega|_{\phi} = |\phi_{t}-\phi_{t-1}| \le \epsilon_{\phi}.
\end{equation}
In practice, we restrict the new ST-glimpse to be within the 8-adjacency of the previous ST-glimpse in the spherical coordinates.
The problem can be reduced to a shortest path problem and solved efficiently using dynamic programming.
The algorithm generates $K$ NFOV outputs from each $360\degree$ input by 1) computing the best trajectory ending at each ST-glimpse location (of $18 \times 11=198$ possible),
and 2) picking the top $K$ of these.
The algorithm assumes the input consists of pixel values on the unit sphere around the camera optical center and does not assume a specific $360\degree$ camera model.
Our results include video frames from at least four common $360\degree$ camera models.

%% file: zoom_lens.tex
\begin{figure}[t]
    \center
    \begin{subfigure}{\linewidth}
        \includegraphics[width=.24\linewidth]{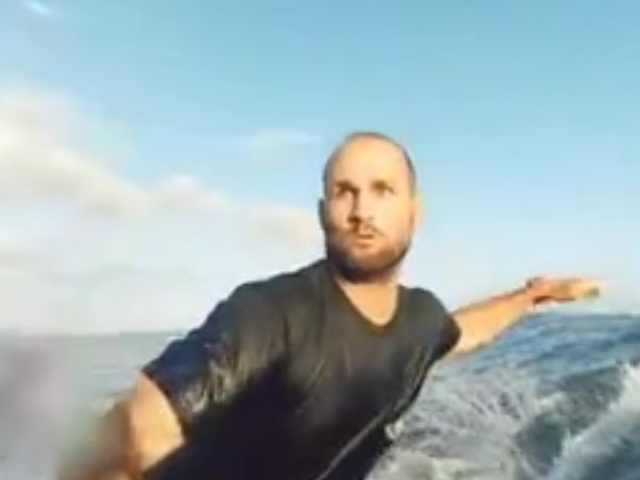}
        \includegraphics[width=.24\linewidth]{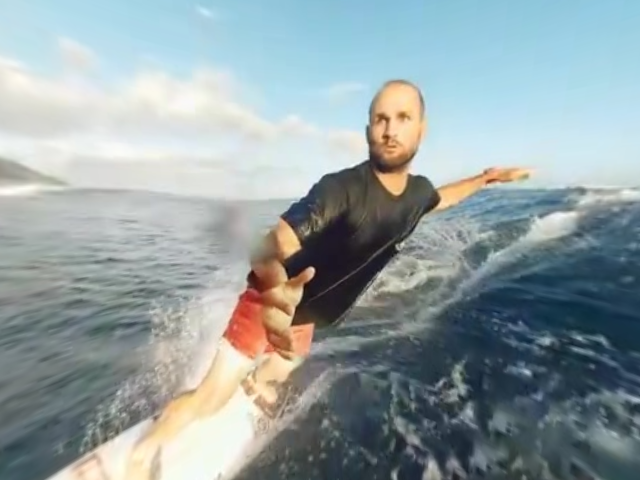}
        \includegraphics[width=.24\linewidth]{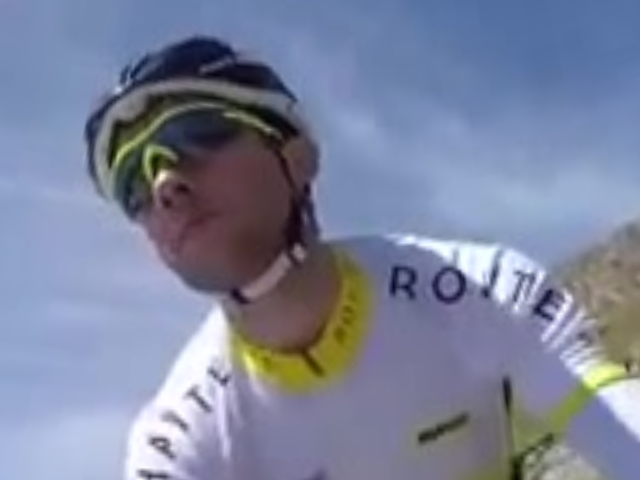}
        \includegraphics[width=.24\linewidth]{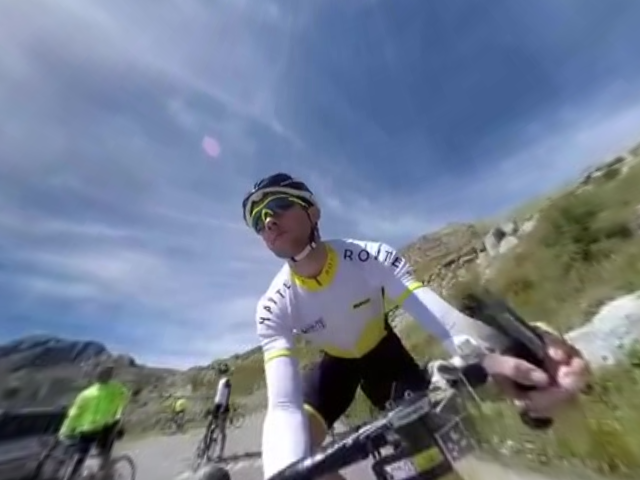}
        \vspace{-4pt}
        \caption{Zoom out helps to capture complete content.}
        \label{fig:qualitative_zoomout}
    \end{subfigure}
    \begin{subfigure}{\linewidth}
        \includegraphics[width=.24\linewidth]{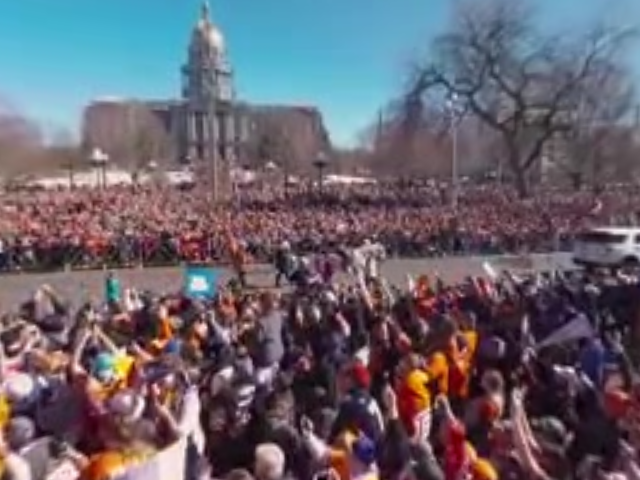}
        \includegraphics[width=.24\linewidth]{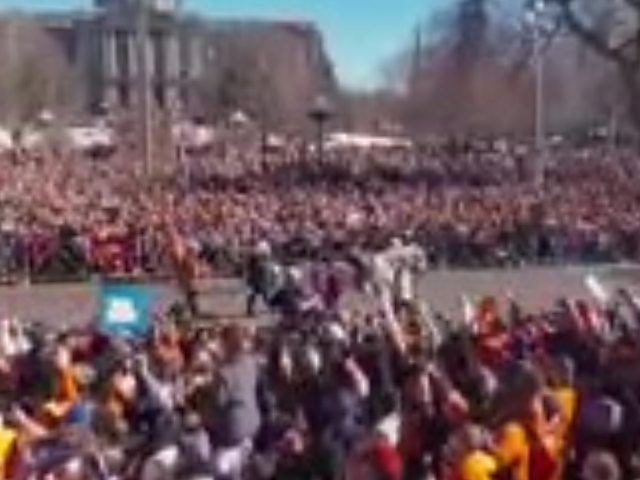}
        \includegraphics[width=.24\linewidth]{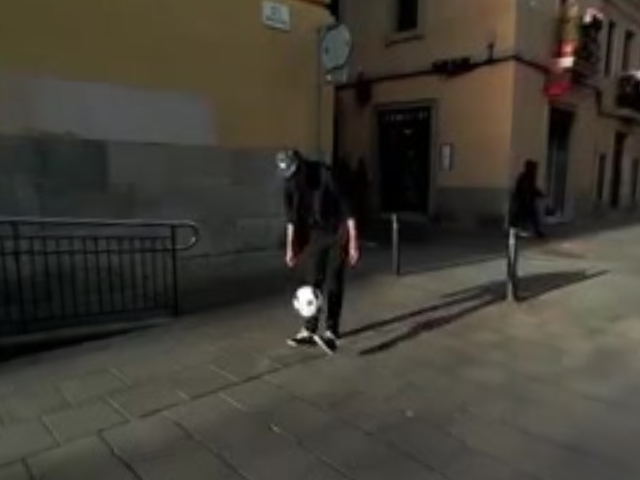}
        \includegraphics[width=.24\linewidth]{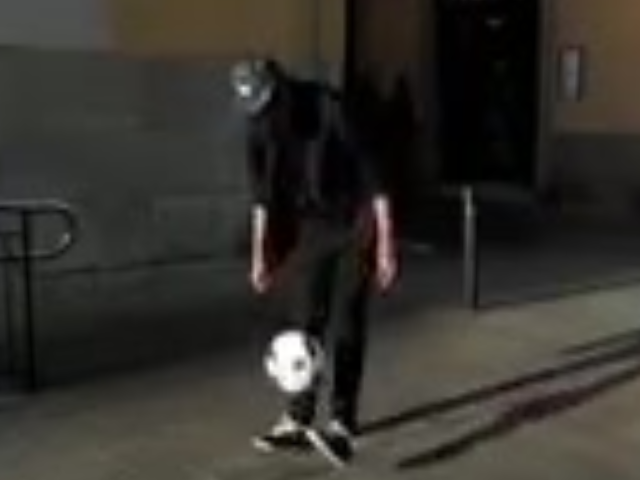}
        \vspace{-4pt}
        \caption{Zoom in helps to concentrate on the subject.}
        \label{fig:qualitative_zoomin}
    \end{subfigure}
    \vspace{-8pt}
    \caption{
        Examples showing the importance of zooming.
        Our zoom in/out result is on the right for each pair.
    }
    \vspace{-12pt}
    \label{fig:qualitative_zoom}
\end{figure}

\begin{figure*}[t]
    \vspace{-9pt}
    \begin{minipage}[b]{.32\textwidth}
        \center
        \includegraphics[width=\textwidth]{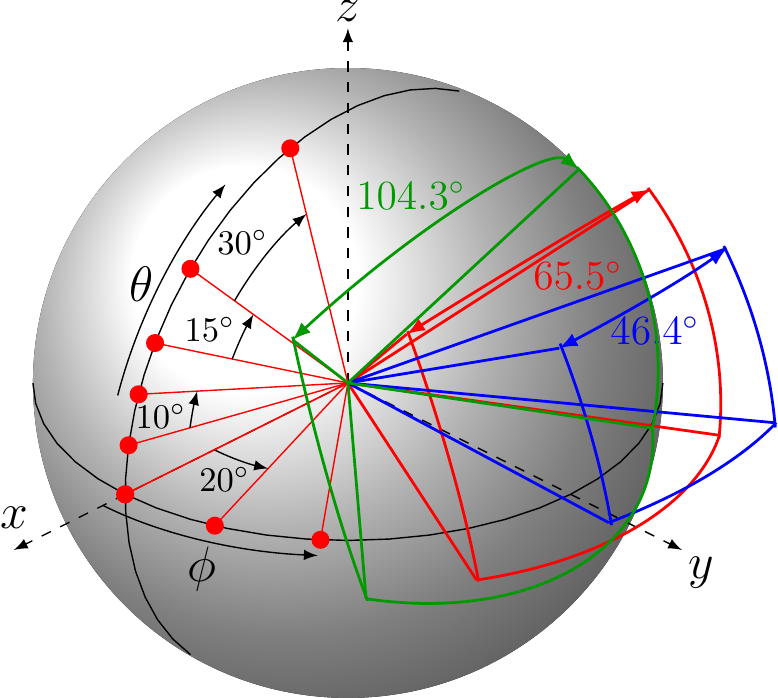}
        \vspace{-18pt}
        \caption[Caption]{Generalized ST-glimpses.\footnotemark}
        \label{fig:generalized_samples}
    \end{minipage}
    \begin{minipage}[b]{.67\textwidth}
        \center
        \includegraphics[width=\textwidth]{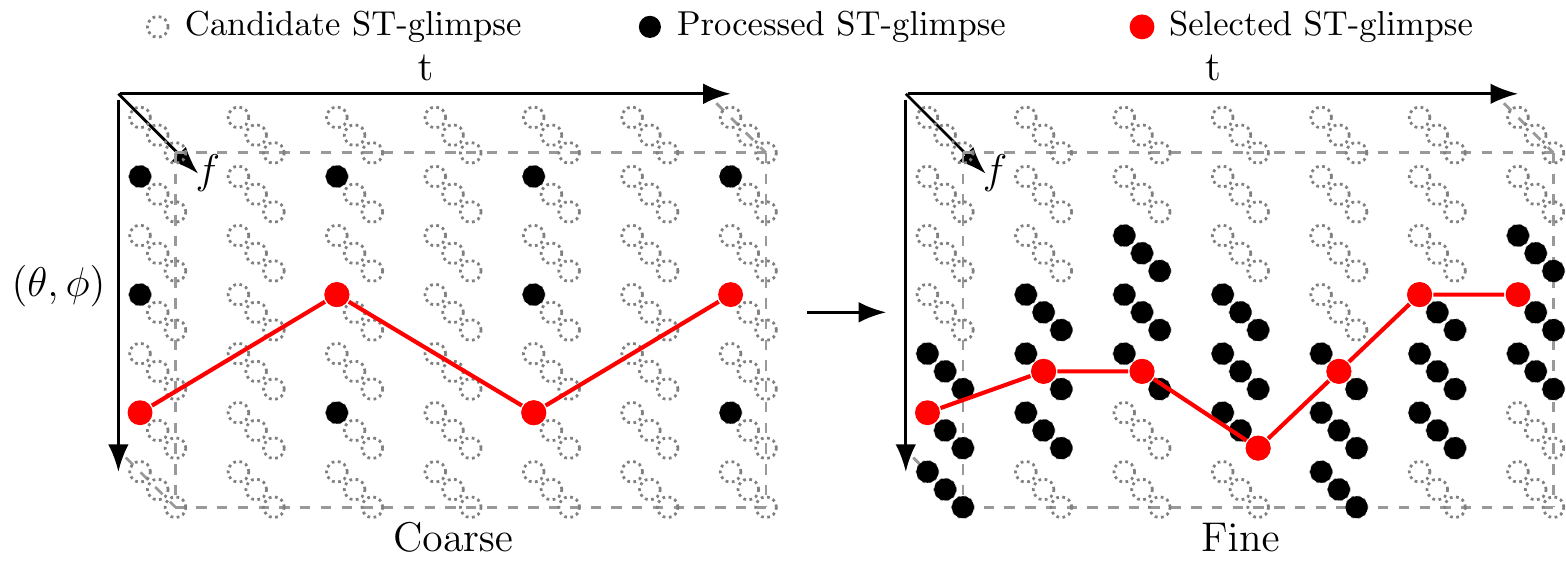}
        \vspace{-21pt}
        \caption{
            Coarse-to-fine camera trajectory search.
            We first construct the trajectory on a coarse sample of ST-glimpses and then refine it on a dense sample of ST-glimpses around the trajectory.
            It reduces computational cost by avoiding processing all candidate ST-glimpses.
        }
        \label{fig:coarse2fine}
    \end{minipage}
    \vspace{-12pt}
\end{figure*}

We first generalize the Pano2Vid problem to Zoom Lens Pano2Vid, which allows zooming in the virtual camera control.
Zooming is the technique of changing the focal length ($f$) of the lens.
This is equivalent to changing the FOV of the camera because they are related by
\begin{equation}
    \text{FOV} = 2 \arctan(\frac{d}{2f}),
\end{equation}
where $d$ is the horizontal sensor size and is a constant for the camera.  The technique is widely used in videography:
almost every digital camera nowadays provides the functionality, and most of us have the experience of adjusting the camera FOV by zooming to obtain the desirable framing.
On the other hand, many existing manual interfaces for viewing $360\degree$ videos do not provide the control for FOV.
Instead of trying to comply with these interfaces,
we believe a proper definition of Pano2Vid must follow the full experience of videography,
i.e.~generating a video as if it is captured by a human videographer in the scene.
Fig.~\ref{fig:qualitative_zoom} shows examples where zooming helps to improve the framing.

To achieve the effect of zooming,
we sample ST-glimpses not only with different principal axis directions but also with different focal lengths.
Assume the focal length for the $65.5\degree$ FOV is $f^{0}$,
we sample ST-glimpses with three different focal lengths
\begin{equation}
    \label{eq:zoom_samples}
    f\,{\in}\,F\,{=}\,\{0.5f^{0},f^{0},1.5f^{0}\},
\end{equation}
which results in $\text{FOV}{\in}\{104.3\degree, 65.5\degree, 46.4\degree\}$ respectively.
The $104.3\degree$ FOV corresponds to an ultra wide angle lens and is the largest FOV commonly used in photography.
The $65.5\degree$ and $46.4\degree$ FOV cover the range of standard lenses.
The sampling for the principal axis direction is kept the same as Eq.~\ref{eq:sample_glimpses}.
The new ST-glimpses are therefore defined by:
\begin{equation}
    \Omega_{t, \theta, \phi, f} \equiv (\theta_t,\phi_t,f_{t}) \in \Theta \times \Phi \times F.
\end{equation}
See Fig.~\ref{fig:generalized_samples}.

When constructing the camera trajectories,
we allow the algorithm to select ST-glimpses with different focal lengths.
Therefore, the problem becomes finding a path over a series of 3D grids ($\Theta \times \Phi \times F$).
In addition to the smoothness constraints in Eq.~\ref{eq:autocam_constraints},
we restrict the change in focal length between consecutive ST-glimpses:
\begin{equation}
    \label{eq:zoom_constraints}
    |\Delta \Omega|_{f} = |f_{t}-f_{t-1}| \le 0.5f^{0},
\end{equation}
which is a prior saying that human videographers tend to use gradual changes in zoom.
While a basic dynamic programming problem similar to that used in \textsc{AutoCam} can find the trajectories in our formulation,
the computational cost grows linearly with respect to the number of available focal lengths and becomes prohibitive.
To make the algorithm for solving the Zoom Lens Pano2Vid problem practical,
we next introduce a new computationally efficient approach for optimizing the trajectories.\footnotetext{Figures are best viewed in color.}

%% file: two_stages.tex
The computational bottleneck for the algorithm is estimating the capture-worthiness score of each ST-glimpse.
The algorithm has to first render the $360\degree$ ST-glimpse into NFOV video and then extract the C3D feature,
both of which are computationally intensive.
A basic dynamic programming approach like \textsc{AutoCam} requires the capture-worthiness scores for all candidate ST-glimpses.
Even if we assume we can render the NFOV video and extract C3D features in real time,
the processing time would be orders of magnitude longer than the input video length due to the large number of candidate ST-glimpses,
i.e.~several hours to process even a 1 minute $360\degree$ video.
Therefore, we aim to reduce the number of capture-worthiness scores required.
The most straightforward way would be to downsample the number of candidate ST-glimpses.
However, this would lead to coarser camera control and a degradation in the quality of the output videos.
In other words,
the computational overhead not only makes the algorithm slow but it also restricts the granularity of virtual camera control.

Instead, we propose a coarse-to-fine approach that preserves the effective number of candidate ST-glimpses while reducing the number of ST-glimpses that require computation of the capture-worthiness score.
The basic idea is to first construct the trajectory over coarsely sampled ST-glimpses and then refine the solution over densely sampled ST-glimpses centered around the initial trajectory.
See Fig.~\ref{fig:coarse2fine}.
Furthermore, to keep the total cost sublinear in the number of available focal lengths,
we construct the coarse trajectory with a single focal length and enable zooming only when refining the solution.
Because we process ST-glimpses densely only in a small portion of the video,
the total number of capture-worthiness scores required by the algorithm decreases.
The proposed coarse-to-fine approach is based on the observation that the capture-worthiness scores of neighbor ST-glimpses are positively correlated,
and the optimal trajectory in densely sampled ST-glimpses leads to a candidate solution in coarsely sampled ST-glimpses.
Although the solution is not guaranteed to be the same as that of dynamic programming over all candidate ST-glimpses,
empirical results verify that it perform well.

We start the algorithm by sampling ST-glimpses at
\begin{equation}
    \label{eq:coarse_glimpses}
    \begin{aligned}
        \theta &\in& \Theta^{\prime} &= \{\pm10,\pm30,\pm75\},\\
        \phi   &\in& \Phi^{\prime}   &= \{0,40,\ldots,320\},\\
        t      &\in& T^{\prime}      &= \{0s,10s,\ldots,L-10s\},\\
        f      &\in& F^{\prime}      &= \{0.5f^{0}\}.
    \end{aligned}
\end{equation}
We use the focal length $f\,{=}\,0.5f^{0}$ which corresponds to the largest FOV so the visual content of these initial ST-glimpses cover that of other focal lengths at the same direction.
Eq.~\ref{eq:coarse_glimpses} downsamples the candidate ST-glimpses by a factor of two from Eq.~\ref{eq:sample_glimpses},
so the number of capture-worthiness scores that need to be computed is only
\begin{equation}
    \frac{|\Theta^{\prime} \times \Phi^{\prime} \times T^{\prime}|}{|\Theta \times \Phi \times T|} \times \frac{1}{|F|} \approx 4.5\%
\end{equation}
the number of candidate ST-glimpses.
We solve for the trajectory using dynamic programming,
but set the smoothness constraint to $2\epsilon$ to account for the coarser samples
\begin{equation}
    \label{eq:coarse_constraints}
    |\Delta \Omega|_{\theta} = |\theta_{t}-\theta_{t-1}| \le 2\epsilon, \;
    |\Delta \Omega|_{\phi} = |\phi_{t}-\phi_{t-1}| \le 2\epsilon.
\end{equation}
Denoting the ST-glimpses selected by the trajectory as
\begin{equation}
    \Omega^{0}_{t,\theta,\phi} \equiv (\theta^{0}_{t},\phi^{0}_{t})
\end{equation}
for $t \in T^{\prime}$,
we then interpolate the ST-glimpses for $t \in T \setminus T^{\prime} = \{5s,15s,\ldots\}$ to obtain the full trajectory.

To refine the trajectory, we sample ST-glimpses
\begin{equation}
    \Omega^{1}_{t,\theta,\phi,f}=(\theta^{1}_{t},\phi^{1}_{t},f^{1}_{t})
\end{equation}
that are adjacent to the original trajectory in direction
\begin{equation}
    \label{eq:neighbor_constraints}
    |\theta^{1}_{t} - \theta^{0}_{t}| \le \epsilon_{\theta}, \; |\phi^{1}_{t} - \phi^{0}_{t}| \le \epsilon_{\phi}
\end{equation}
following Eq.~\ref{eq:sample_glimpses} and \ref{eq:zoom_samples}.
We then solve the same trajectory search problem over the sampled ST-glimpses using dynamic programming with the smoothness constraints in Eq.~\ref{eq:autocam_constraints} and Eq.~\ref{eq:zoom_constraints}.
The number of candidate ST-glimpses is greatly reduced by the adjacency constraint in Eq.~\ref{eq:neighbor_constraints} and is only $5\%$ of all candidate ST-glimpses.

%% file: diverse_search.tex
\begin{figure}[t]
    \hspace{-12pt}
    \vspace{-2pt}
    \includegraphics[width=1.06\linewidth]{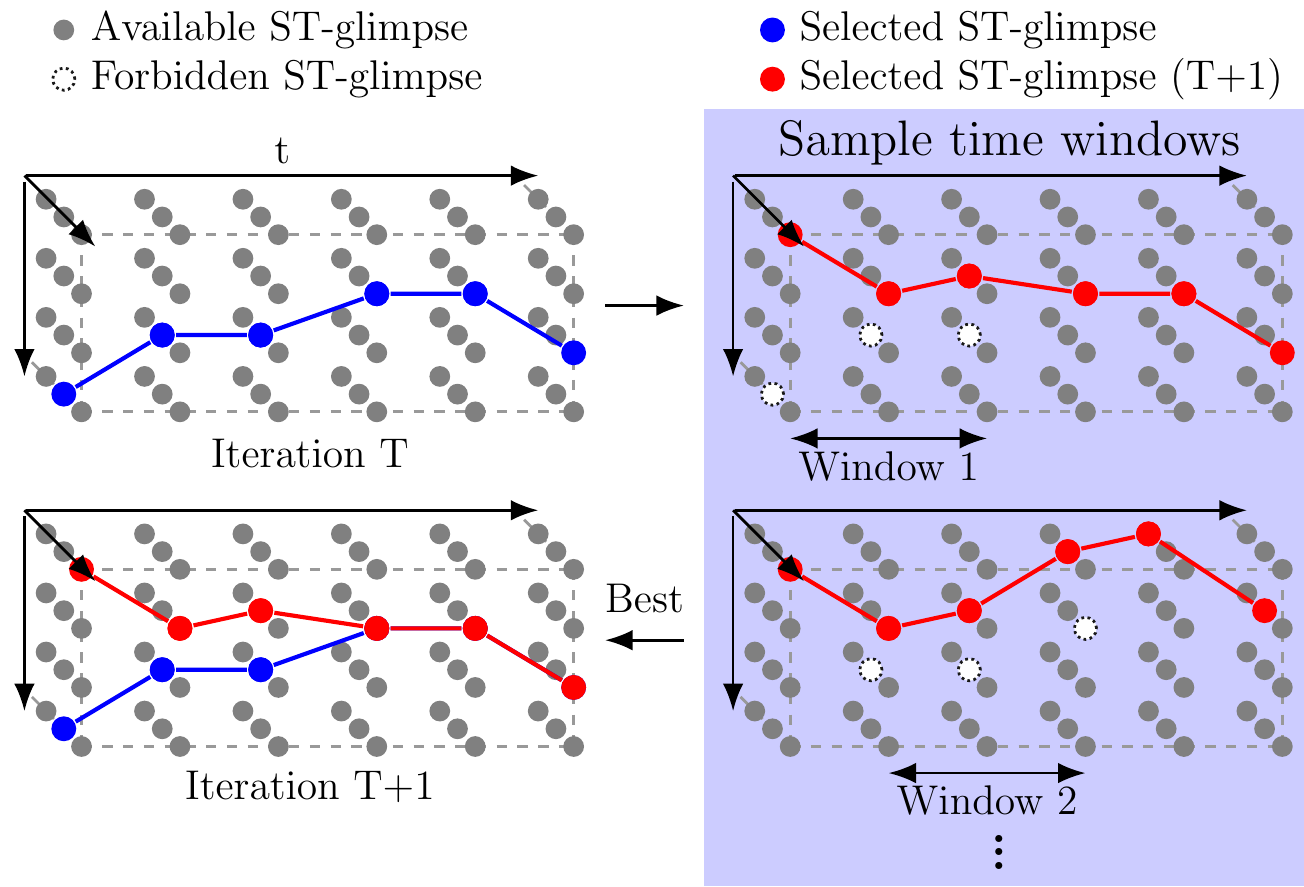}
    \vspace{-18pt}
    \caption{
        Diverse trajectory search generates trajectories iteratively.
        In each iteration, we construct multiple trajectory search problems by sampling time windows and removing previously selected ST-glimpses in the window from the search space.
        We solve all the problems and take the best solution as the output trajectory.
    }
    \vspace{-12pt}
    \label{fig:diverse}
\end{figure}

So far we have 1) presented our approach for the generalized Pano2Vid problem allowing variable FOV, and 2) devised a fast algorithm for optimizing it to produce the best hypothesis NFOV video output.  Next, we wish to expand that single best solution to a set of \emph{diverse} plausible outputs.

The motivation to generate a diverse set of output videos stems from the fact that, by definition, there may be multiple valid Pano2Vid solutions for each $360\degree$ video.
For example, one might capture a soccer game by tracking the ball or by focusing on a particular player.
Both of them will lead to a plausible presentation for the game and should be a valid output.
In fact, when we ask human editors to manually extract NFOV videos from $360\degree$ data,
the outputs for any two editors on the same source video have only about 47\% overlap on average.
In addition, many applications of Pano2Vid would prefer a set of candidate solutions instead of a single output.
An editing aid system would be more useful if the editor has the freedom to choose from different reasonable algorithm-provided initializations.
Similarly, a $360\degree$ video player that allows the viewers to choose from different NFOV video presentations is likely to achieve a better user experience because the viewers can decide what to see based on their preferences.

It is difficult to encourage diversity in a single pass of dynamic programming,
because all the potential solutions are constructed concurrently and the distances between them are hard to control.
In prior work, the \textsc{AutoCam} algorithm generates multiple trajectories by requiring them to end at different spatial locations in the video,
i.e., with different ST-glimpses.
However, this requirement does not guarantee the distance between different trajectories.
In the worst case, they can be exactly the same except in the last ST-glimpses, leading to poor diversity in the outputs.

Instead, we generate trajectories iteratively and encourage diversity by imposing the minimum distance constraint between trajectories generated in different iterations.
To realize the constraint,
we sample a time window and forbid the trajectories of the current iteration from selecting the same ST-glimpses as the solution of previous iterations in the window.
Therefore, the length of the time window determines the minimum distance between the solutions of different iterations.
We sample the time window at multiple temporal locations and construct the optimal trajectory for each window.
We take the best trajectory among them in terms of accumulated capture-worthiness score as the solution of the current iteration.
This avoids critical glimpses being excluded from the solution space even if it is selected by previous trajectories.
See Fig.~\ref{fig:diverse}.

In practice, we set the length of the time window to $10\%$ the original video length and sample 20 different windows distributed evenly over time.
Once the length and location of the time window is specified,
the optimal trajectory can be found using dynamic programming on a modified shortest path problem where the ST-glimpses selected by previous solutions in the window are removed from the search space.
To improve computational efficiency,
we divide the unit sphere into 6 regions (by 3 azimuthal angles and 2 polar angles) and generate an output per region by finding the best trajectory ending in the region.
This leads to 6 trajectories per iteration.
The iteration ends after $K$-trajectories have been generated.

%% file: dataset.tex
We use the Pano2Vid dataset introduced in~\cite{su2016accv}.\footnote{\url{http://vision.cs.utexas.edu/projects/Pano2Vid}}
It consists of 86  $360\degree$ videos crawled from YouTube using four keywords: ``Soccer,'' ``Mountain Climbing,'' ``Parade,'' and ``Hiking", for a total length of 7.3 hours.
The dataset also provides ``HumanCam" data: URLs for 343 hours of NFOV YouTube videos.  We use half of that data to train logistic regression capture-worthiness classifiers and half for evaluation (details below).
Following~\cite{su2016accv}, we train the classifier using leave-one-$360\degree$-video-out strategy for each keyword.

Some evaluation metrics below compare our outputs to human-selected camera trajectories from $360\degree$ videos (``HumanEdit'').  To collect these trajectories,
we ask human subjects to edit the videos.  The editors control the angle and FOV of a virtual camera with their mouse, overlayed on the $360\degree$ video's equirectangular projection so the user can see all the visual content at once.
Please see supp.~file for details and examples.
We collect  HumanEdit data for 40 videos, each of them annotated by 3 editors.
Overall, we collect 480 trajectories totaling 717.2 minutes of video, and roughly 18 hours of annotation time.

%% file: baselines.tex
We compare the following methods in the experiments.
\begin{itemize}[leftmargin=*,label=$\bullet$]
    \vspace*{-0.08in}
    \setlength{\itemsep}{2pt}
    \setlength{\parskip}{2pt}
    \item \textsc{Center} --- random trajectories biased toward the ``center'' of the $360\degree$ video axes ($\theta=\phi=0\degree$).
        The bias accounts for the fact that user-generated $360\degree$ videos often contain interesting content close to the center,
        possibly because the $360\degree$ camera design allows the users to use it as if it were a NFOV camera.
         We sample the camera direction for the next time-step from a Gaussian centered around the current direction, which starts from the center.
    \item \textsc{Eye-level} --- static trajectories that place the virtual camera on the equator ($\theta=0\degree$).
        The equator usually corresponds to eye-level in $360\degree$ video where most interesting events happen.
        We sample the azimuthal angle $\phi$ every $20\degree$ for 18 different camera directions.
    \item \textsc{Saliency} --- replace the capture-worthiness scores in \textsc{AutoCam} by saliency scores.\footnote{We also considered a saliency baseline that permits zoom like our method, but it fared worse than all others.} The saliency is computed by a popular method~\cite{harel2006graph} over the $360\degree$ video frame in equirectangular projection.
    \item \textsc{AutoCam} --- to our knowledge, the only prior work tackling this problem~\cite{su2016accv}.
\end{itemize}

%% file: metrics.tex
\begin{table*}[t]\scriptsize
    \setlength\extrarowheight{2pt}
    \vspace{-9pt}
    \centering
    \caption{
        Pano2Vid performance: HumanCam-based and HumanEdit-based metrics.
        The arrows in column 3 indicate whether lower scores are better ($\Downarrow$), or higher scores ($\Uparrow$).
        Our full method (\textsc{Ours}) significantly outperforms the baselines, and the relative improvements over the best performing baseline are up to $43.4\%$.
    }
    \vspace{-24pt}
    \label{tab:quality}
    \resizebox{\textwidth}{!}{
    \begin{tabular}{llccccc|cc}
        & & & \multirow{2}{*}{\textsc{Center}} & \multirow{2}{*}{\textsc{Eye-level}} & \multirow{2}{*}{\textsc{Saliency}} & \multirow{2}{*}{\textsc{AutoCam}~\cite{su2016accv}} & \textsc{Ours}          & \multirow{2}{*}{\textsc{Ours}} \\[-2pt]
        & & &                                  &                                     &                                    &                                   & \textsc{w/o Diversity} & \\[-0.8pt]
        \hline
        Distinguishability & Error rate (\%) & $\Uparrow$   & 1.93 & 4.03 & 7.70 & 12.05 & \textbf{17.41}  & 17.28 \\
        \hline
        HumanCam-Likeness  & Mean Rank       & $\Downarrow$ & 0.659 & 0.707 & 0.612 & 0.522 & 0.279 & \textbf{0.267}\\
        \hline
        \multirow{2}{*}{Transferability} & Human $\rightarrow$ Auto & \multirow{2}{*}{$\Uparrow$} & 0.582 & \textbf{0.607} & 0.597 & 0.517 & 0.590          & 0.591 \\
                                         & Auto $\rightarrow$ Human &                             & 0.526 & 0.552          & 0.549 & 0.584 & \textbf{0.618} & 0.617 \\
        \hline
        \multirow{2}{*}{Overlap} & Trajectory & \multirow{2}{*}{$\Uparrow$} & 0.271 & 0.335 & 0.359 & 0.343 & 0.436 & \textbf{0.442} \\
                                 & Frame      &                             & 0.498 & 0.555 & 0.580 & 0.530 & 0.629 & \textbf{0.630} \\
        \noalign{\hrule height .6pt}
    \end{tabular}
    }
    \vspace{-18pt}
\end{table*}

We adopt the metrics proposed in~\cite{su2016accv}, generalizing them as need to account for Zoom Lens Pano2Vid.

\vspace*{-0.16in}
\paragraph{HumanCam-based metrics}

This group of metrics evaluates whether the output videos look like human-captured NFOV video (HumanCam).
The more indistinguishable the algorithm outputs are from HumanCam, the better the algorithm.
There are three metrics:
\begin{itemize}[leftmargin=*,label=$\bullet$]
    \vspace*{-0.06in}
    \setlength{\itemsep}{2pt}
    \setlength{\parskip}{2pt}
    \item \textbf{Distinguishability} quantifies if it is possible to tell the algorithm output apart from HumanCam.
        It is measured by 5-fold cross validation error rate of discriminative classifiers trained with HumanCam as positives and algorithm outputs as negatives.
        \emph{Higher} error is better.
    \item \textbf{HumanCam-likeness} measures the relative distance from algorithm outputs to HumanCam data in a semantic feature space.
        It trains classifiers with HumanCam video as positives and all algorithm-generated videos as negatives using leave-one-$360\degree$-video-out strategy.
        An algorithm-generated video is ranked based on its distance to the decision boundary,
        and we compute the normalized mean rank; lower is better.
    \item \textbf{Transferability} measures how well semantic classifiers trained on HumanCam videos transfer to algorithm-generated videos, and vice versa.
        The more transferable the classifiers are, the more similar the HumanCam and algorithm outputs are.
        We take the four search keywords as labels to learn a multi-class classifier on one domain and measure transferability using the test error on the other domain.
\end{itemize}
\vspace*{-0.1in}
We use logistic regression classifiers in all metrics.

\begin{figure}[t]
    \center
    \includegraphics[width=\linewidth]{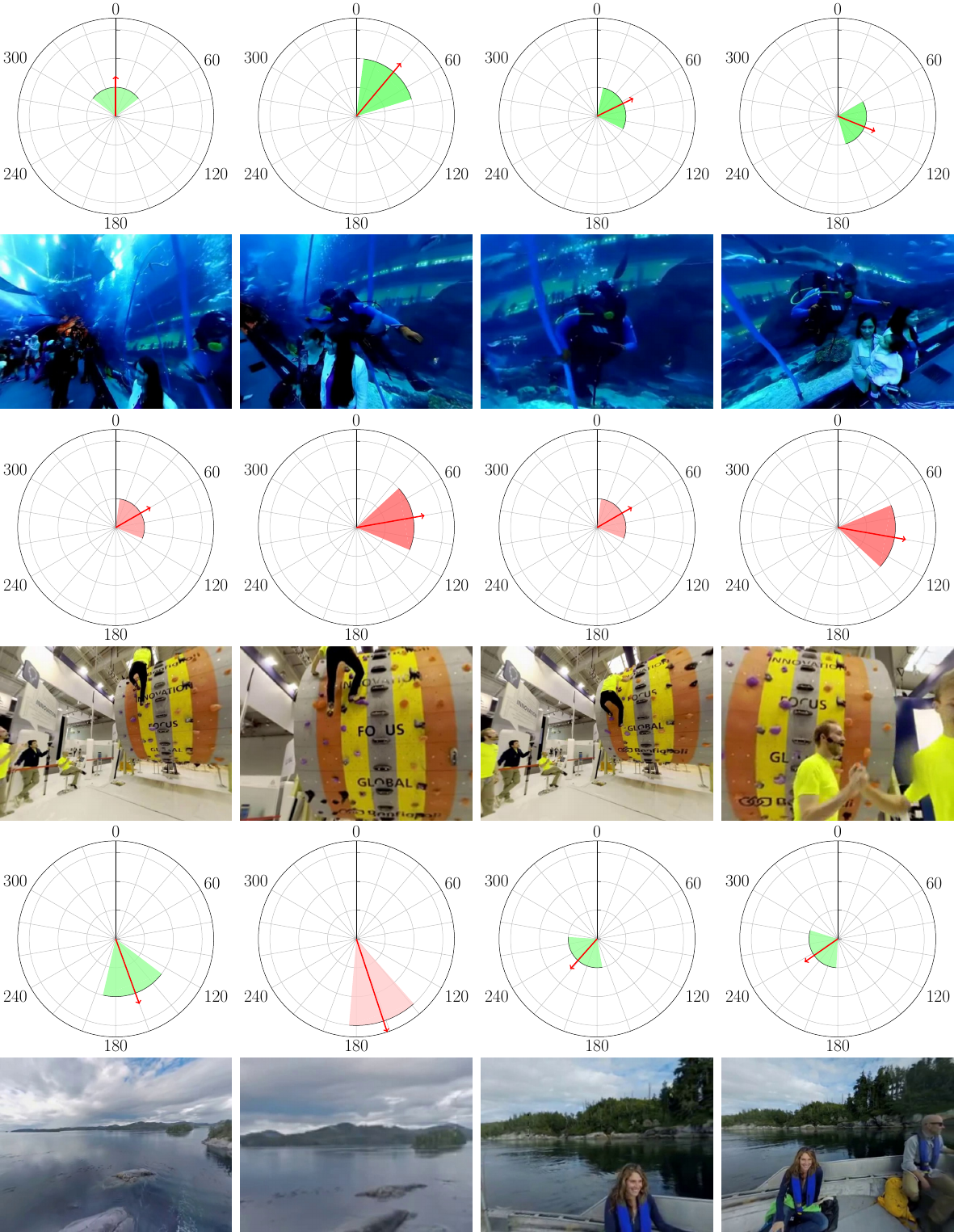}
    \vspace{-18pt}
    \caption{
        Example frames of our algorithm outputs and the corresponding camera poses.
        The circular sector shows the camera FOV and azimuthal angle, and the color shows the polar angle.
        Red/green indicates the angle is greater/smaller than 0,
        and more saturated color indicates larger value.
        In the first example, the camera tracks the diver.
        In the second example, it zooms in to capture special moments,
        e.g., when the person climbs to the top or high fives with another person.
        In the third example, the camera first captures distant scene with long focal length and close objects with short focal length.
        Also see videos on the project webpage.
    }
    \vspace{-12pt}
    \label{fig:qualitative1}
\end{figure}

\begin{figure}[t]
    \center
    \includegraphics[width=\linewidth]{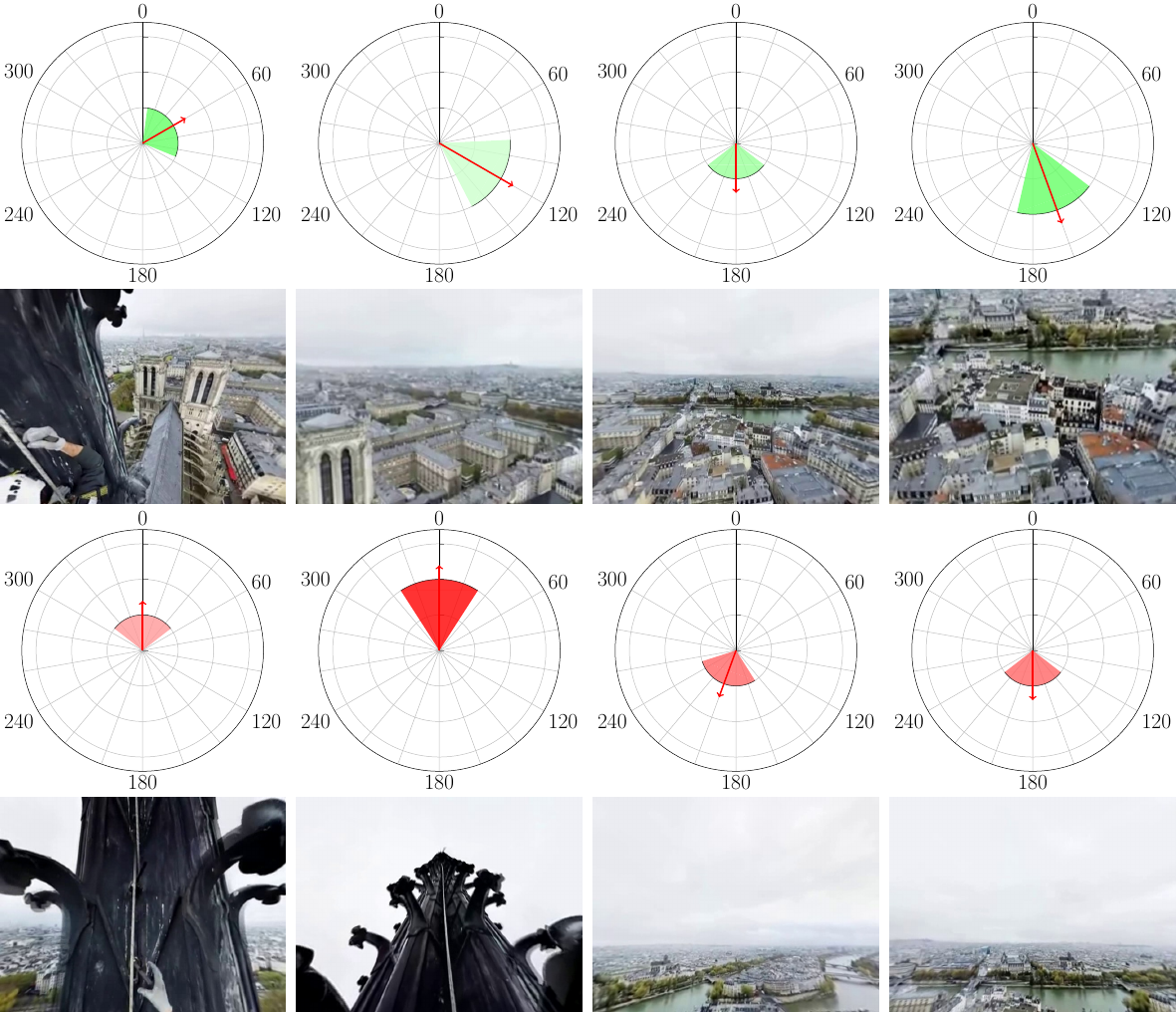}
    \vspace{-18pt}
    \caption{
        Two trajectories extracted from the same $360\degree$ video.
        Our algorithm presents the same scene in different manners.
    }
    \label{fig:qualitative2}
    \vspace{-12pt}
\end{figure}

\vspace*{-0.12in}
\paragraph{HumanEdit-based metric}

Whereas the above metrics score output videos by their resemblance to human-captured videos in general,
the HumanEdit metric measures the similarity between algorithm-generated camera trajectories and the manually created trajectories in the same $360\degree$ video.
The more similar they are, the better the algorithm.
This metric captures the subjective preference of human editors but is easily reproducible,
in contrast to one-off user studies.
In particular, we compute the \textbf{overlap} of the camera FOV in each frame.
The overlap is approximated by max($1-\frac{2\Delta \Omega}{FOV_{H}+FOV_{A}},0$),
where $FOV_{H}$ and $FOV_{A}$ correspond to the FOV of algorithm and human controlled camera respectively.
We report overlap results under two pooling strategies,
\textbf{trajectory}, which rewards outputs similar to at least one HumanEdit trajectory as a whole,
and \textbf{frame}, which rewards outputs similar to any HumanEdit trajectory at each frame.  See supp.~for details.

\vspace*{-0.15in}
\paragraph{Diversity}

We gauge diversity by the number of distinct groups of trajectories among the outputs, where trajectories differing in angle or FOV for fewer than $10\%$ of the video frames are considered to be the same group.

\vspace*{-0.15in}
\paragraph{Computational cost}

We measure computational cost by the average processing time for 1 minute of $360\degree$ video.
The time is measured on a machine equipped with one Intel Xeon E5-2697 v2 processor (24 cores) and one GeForce GTX Titan Black GPU (including I/O).

%% file: output_quality.tex
First we quantify the quality of the algorithm-generated videos using the HumanCam-/HumanEdit-based metrics.
We obtain the trajectories for \textsc{AutoCam} from the authors of~\cite{su2016accv}.
We take the top $K{=}20$ outputs, following~\cite{su2016accv}.

Table~\ref{tab:quality} shows the results.
Our full method (\textsc{Ours}) significantly outperforms all other methods.
Compared to the best performing baseline (\textsc{AutoCam}),
our method improves Distinguishability by $43.4\%$ and ranks $25.5\%$ better on average in the HumanCam-Likeness.
We also see a $23\%$ improvement in the Trajectory Overlap metric.
The superior performance clearly shows the importance of the proposed zoom lens.
Figures~\ref{fig:qualitative1} and~\ref{fig:qualitative2} show examples.
Interestingly, the camera zooms out (i.e.~FOV${>}65.5\degree$) more often than it zooms in.
In fact, $76\%$ of the ST-glimpses selected by \textsc{Ours} have $104.3\degree$ FOV,
and editors select $104.3\degree$ FOV in $55\%$ of the HumanEdit data.
These results suggest that a larger FOV is preferable when viewing $360\degree$ videos,
possibly because the object of interest is usually closer to the camera.
Please see project webpage for videos.

Table~\ref{tab:quality} also shows that the \textsc{Center} and \textsc{Eye-level} baselines perform poorly,
indicating that hand coded heuristics based on prior knowledge are not enough and a content-dependent method is necessary.
\textsc{Eye-level} performs better than \textsc{Center},
which reflects the fact that \textsc{Eye-level} is a generic prior while the \textsc{Center} prior only holds when the $360\degree$ camera is asymmetric and the user uses it as an ordinary camera.
Although \textsc{Saliency} is content-dependent,
it captures content that attracts gaze, which appears to be a poor proxy for the Pano2Vid task.
It underperforms \textsc{Ours} in all metrics except the Human $\rightarrow$ Auto transferability.

Although our method outperforms all baselines,
we notice that the learned capture-worthiness is unable to capture preferences induced by context.
For example, the algorithm fails to concentrate on family members in a family video.
Also, the smoothness constraint may be too strong in some scenes where the camera is unable to adjust to rapidly changing content.
See project page for failure cases.

%% file: computational_cost.tex
\begin{figure}[t]
    \vspace{-9pt}
    \center
    \includegraphics[width=0.9\linewidth]{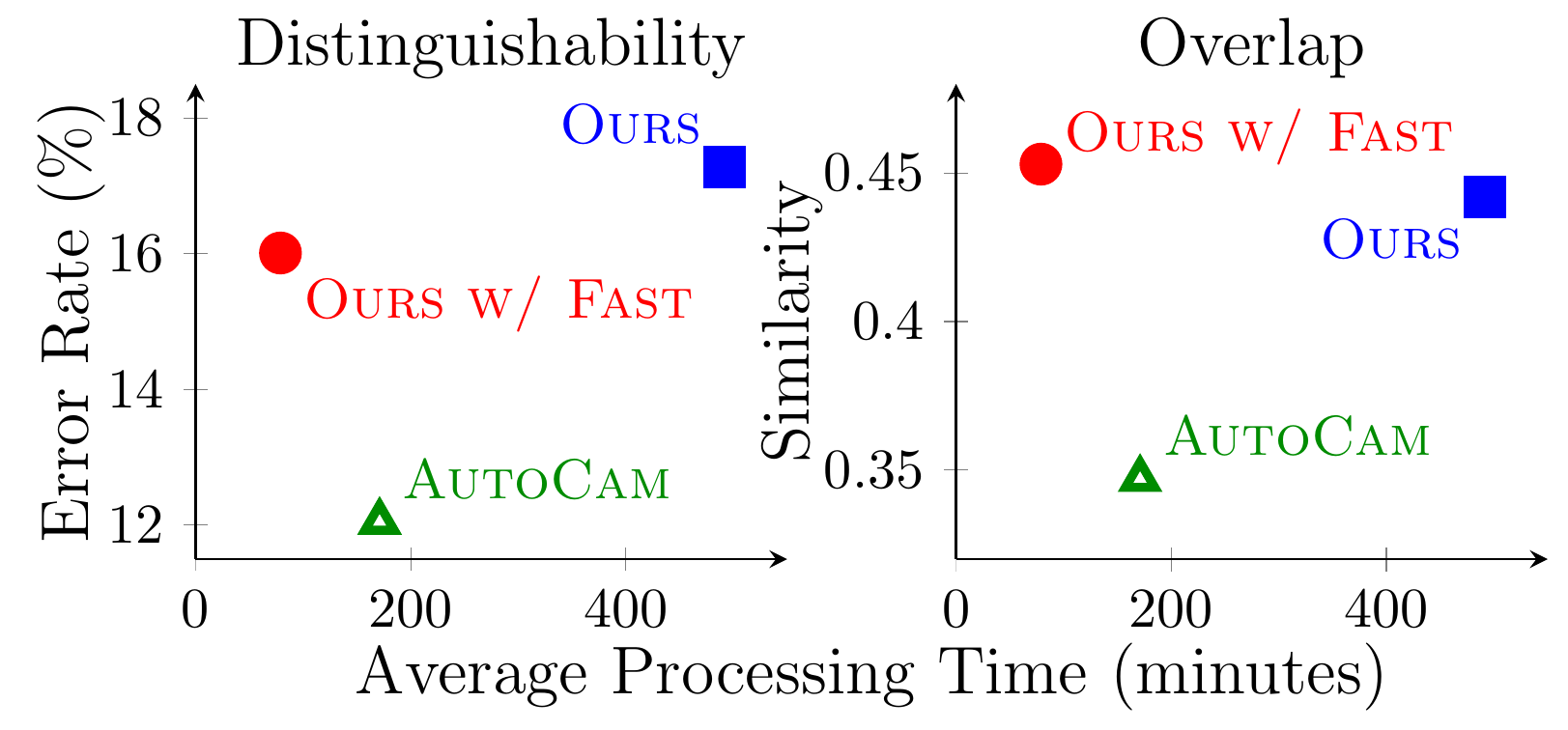}
    \vspace{-9pt}
    \caption{
        Computational cost vs.~output quality for our method and \textsc{AutoCam}~\cite{su2016accv}.
        Computational cost is measured by the average processing time per 1 minute of input $360\degree$ video.
        Quality is measured by the Distinguishability and Trajectory Overlap;
        higher is better ($\Uparrow$) for both metrics.
    }
    \vspace{-12pt}
    \label{fig:computational_cost}
\end{figure}

Fig.~\ref{fig:computational_cost} shows the computational cost versus output quality,
measured by HumanCam Distinguishability and HumanEdit Trajectory Overlap.
The computational cost for \textsc{Ours} \emph{without} coarse-to-fine processing is $2.88$ times that of \textsc{AutoCam},
which shows the need for a more computationally efficient algorithm.
\textsc{Ours w/ Fast} reduces the computational cost by $84\%$ compared to \textsc{Ours} while performing similarly in Trajectory Overlap and only $6\%$ worse in Distinguishability.
Comparing the trajectories generated by the two methods,
the coarse-to-fine approach tends to favor the largest FOV and ignore trajectories that select the smallest FOV throughout the video,
because the initial trajectories are constructed with the largest FOV.
This may cause the degradation in Distinguishability due to the fact that the $104.3\degree$ FOV introduces distortion in output frames.
\textsc{Ours w/ Fast} takes less than $50\%$ of the computation of \textsc{AutoCam} yet is more accurate in all metrics.
Although further optimization on the implementation may reduce the processing time,
the relative cost will remain the same, as it is linear in the number of ST-glimpses processed.

%% file: output_diversity.tex
\begin{figure}[t]
    \vspace{-9pt}
    \center
    \includegraphics[width=0.78\linewidth]{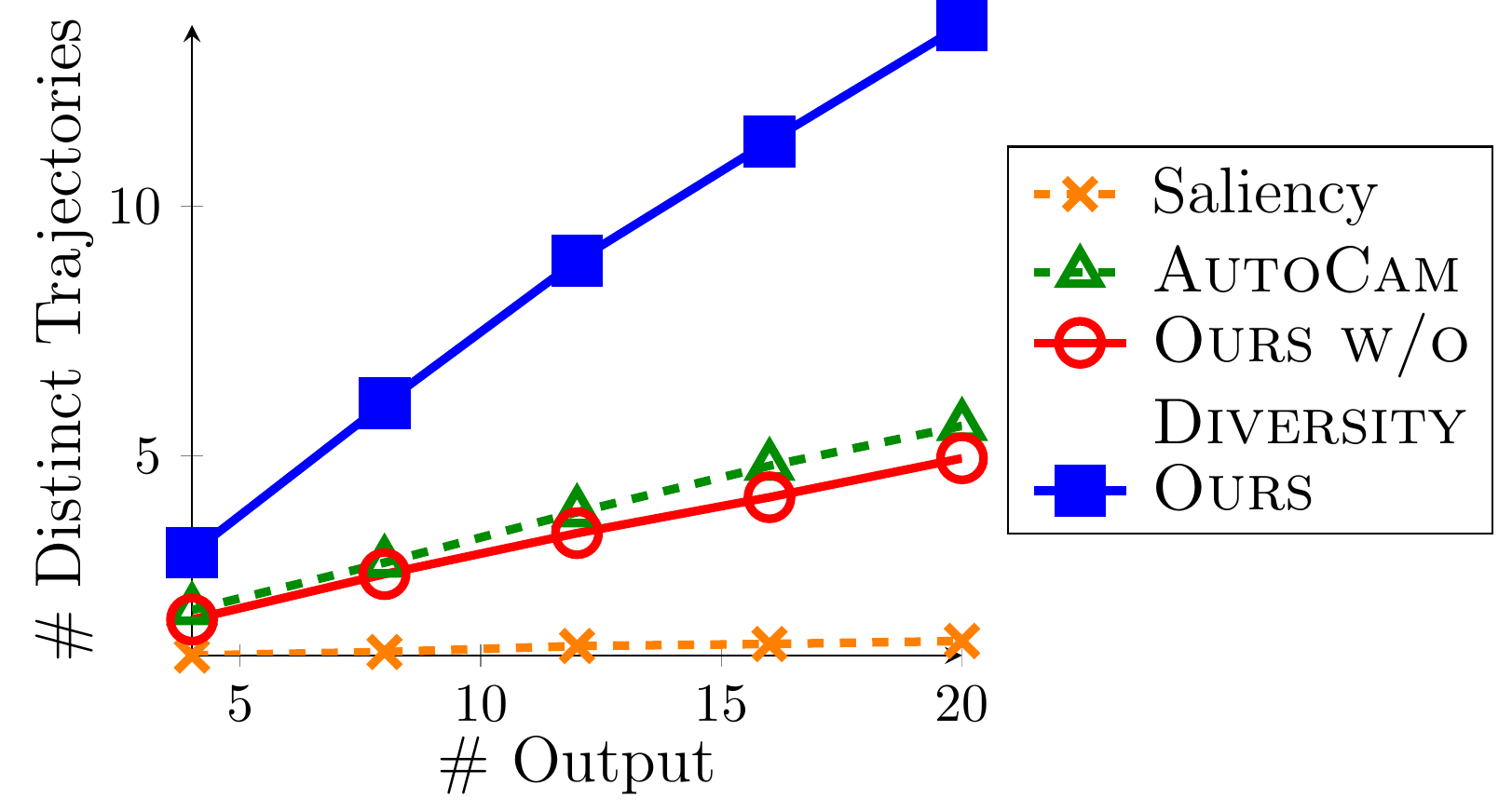}
    \vspace{-9pt}
    \caption{
        The number of distinct trajectories each algorithm generates with respect to the number of algorithm outputs.
    }
    \vspace{-12pt}
    \label{fig:distinct_trajs}
\end{figure}

Fig.~\ref{fig:distinct_trajs} displays the relative diversity captured by the methods as a function of the number of outputs they generate.  
Our diverse trajectory search approach generates more distinct trajectories than any other method, with $2 \sim 3$ times more distinct outputs than our ablated non-diverse variant.
\textsc{AutoCam} generates slightly more distinct trajectories than \textsc{Ours w/o Diversity},
despite the fact that they both use DP to search for the trajectories.
This is because \textsc{Ours w/o Diversity} has more degrees of freedom in camera control and can generate closer variants of a given camera trajectory,
i.e.~it can generate more outputs by varying the choice of the last ST-glimpse than \textsc{AutoCam} can.
\textsc{Saliency} performs poorly in terms of output diversity---fewer than two distinct trajectories on average even among its top 20 outputs.  
Because saliency scores are computed more densely than the capture-worthiness scores,
the correlation of the scores between neighbor directions are stronger,
and the algorithm can generate more trajectories that are close variants to the others.
Note that the results in Fig.~\ref{fig:distinct_trajs} are complementary to that in Table~\ref{tab:quality}.
Together, they show that our method can generate diverse yet quality outputs.

%% file: conclusion.tex
We explore virtual videography in the context of $360\degree$ video.
Our system controls a virtual camera in a $360\degree$ video to generate videos that look human-captured and properly present the content to a passive human viewer.
We generalize the previously proposed Pano2Vid problem by allowing the algorithm to control its field of view dynamically,
introduce a coarse-to-fine optimization approach that makes it tractable,
and propose a method to encourage diversity and capture the multimodal nature of ``good'' NFOV videos.
Under a battery of metrics---including against human editors---our algorithm outperforms the state of the art, bringing tools one step closer to practical virtual videography.

%% file: supp.tex
\section*{Appendix}

The appendix consist of:
\begin{enumerate}[label=\Alph*]
    \item Details for the HumanEdit-based metric
    \item Complete computational cost experiment results
    \item Annotation interface introduction
\end{enumerate}
Please refer to the project webpage at \url{http://vision.cs.utexas.edu/projects/watchable360/} for the annotation interface demonstration and example output videos.

\section{HumanEdit-Based Metric}

We deploy two different strategies for pooling the frame-wise \textbf{overlap}:
\begin{itemize}[leftmargin=*,label=$\bullet$]
    \item \textbf{Trajectory pooling} rewards algorithm outputs that are similar to \emph{at least one} HumanEdit trajectory over the entire video.
        It first computes the per video overlap between each algorithm-generated trajectory and HumanEdit trajectory using the average per frame overlap.
        Each algorithm output is then assigned the score as the overlap with its most similar HumanEdit trajectory.
    \item \textbf{Frame pooling} rewards algorithm outputs that are similar to \emph{any} HumanEdit trajectory at each frame.
        For each algorithm output, it first scores each frame using its overlap to the most similar HumanEdit trajectory.
        The output trajectory is then assigned the score as the average per frame score.
\end{itemize}

\section{Computational Cost Results}

\begin{figure}[t]
    \center
    \includegraphics[width=\linewidth]{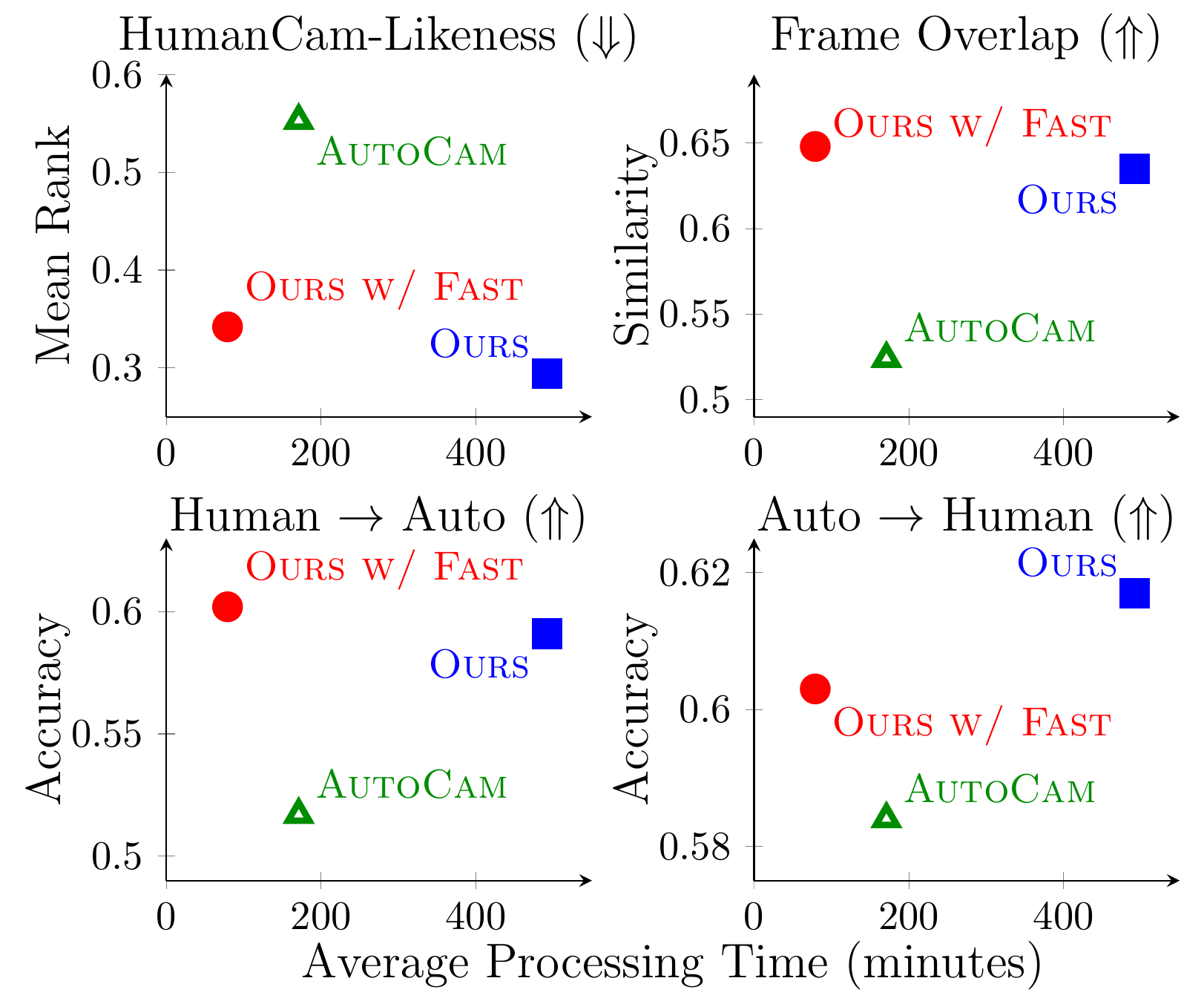}
    \vspace{-18pt}
    \caption{
        Computational cost versus output quality.
        The arrows in title indicate higher scores better ($\Uparrow$) or lower scores better ($\Downarrow$).
        The results are consistent with the distinguishability and trajectory overlap metrics in the main paper,
        and were pushed to supp.~due to space limits.
    }
    \label{fig:computational_cost_supp}
\end{figure}

For completeness,
Fig.~\ref{fig:computational_cost_supp} shows the computational cost versus output quality for all metrics as noted in footenote 6 of the main paper.
Due to space limits,
the main paper includes the same result for Distinguishability and Trajectory overlap in Fig~\ref{fig:computational_cost}.
\textsc{Ours w/ Fast} significantly outperforms \textsc{AutoCam}~\cite{su2016accv} in all metrics.
It performs similarly to \textsc{Ours} in Transferability and Frame Overlap but worse in the HumanCam-Likeness metric.
This is consistent with the Distinguishability metric and is possibly due to the distortion in $104.3\degree$ FOV.
Note the HumanCam-Likeness metric is measured by the normalized ranking and is a relative metric,
so the absolute value depends on the number of methods evaluated and is different from the results in the paper.

\section{HumanEdit Interface}

\begin{figure*}[t]
    \includegraphics[width=\linewidth]{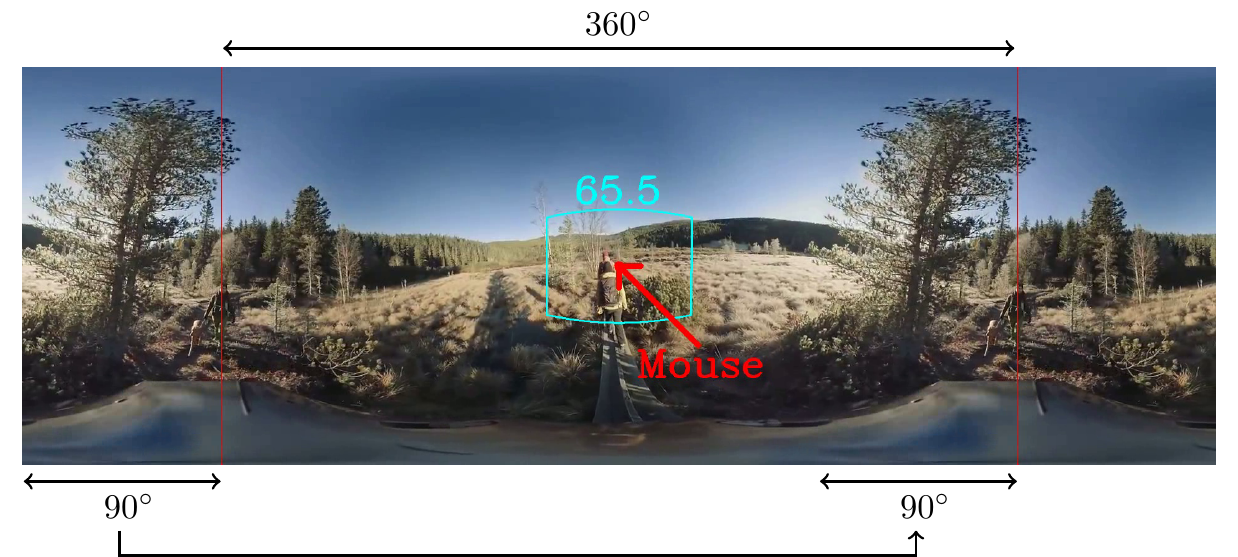}
    \caption{The interface shows the video in panoramic strip. It further expands both side by $90\degree$.}
    \label{fig:interface_angle}
\end{figure*}

\begin{figure*}[t]
    \center
    \begin{subfigure}{\linewidth}
        \includegraphics[width=\linewidth]{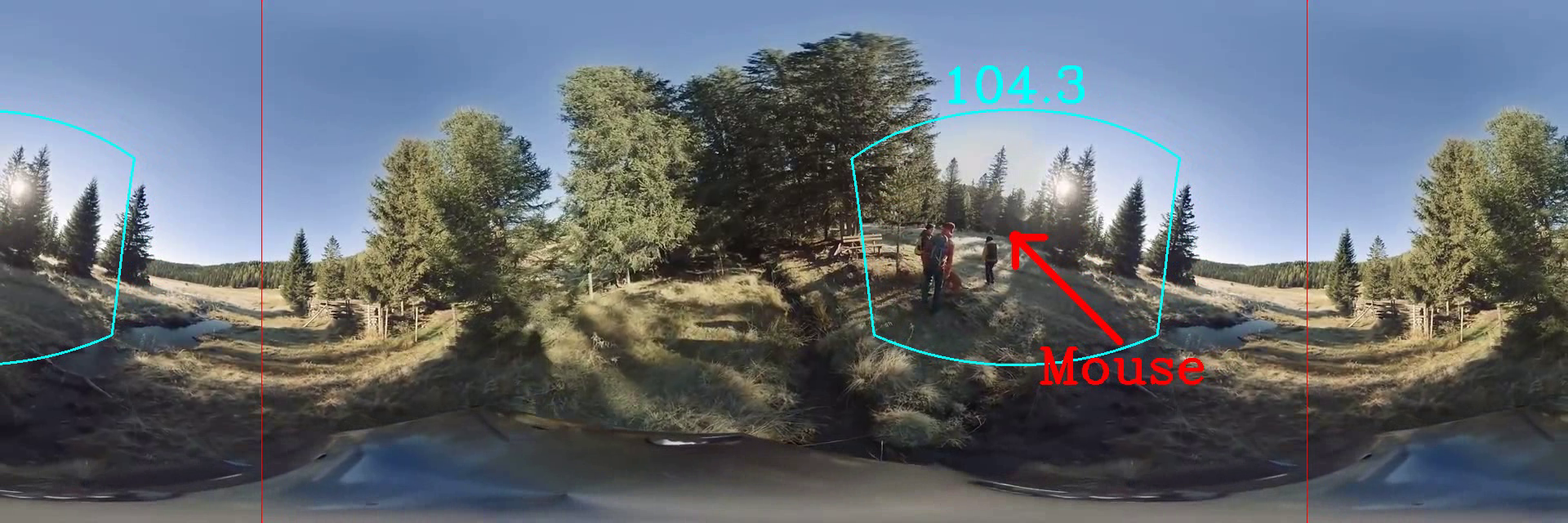}
        \caption{Zoom out.}
        \vspace{8pt}
    \end{subfigure}
    \begin{subfigure}{\linewidth}
        \includegraphics[width=\linewidth]{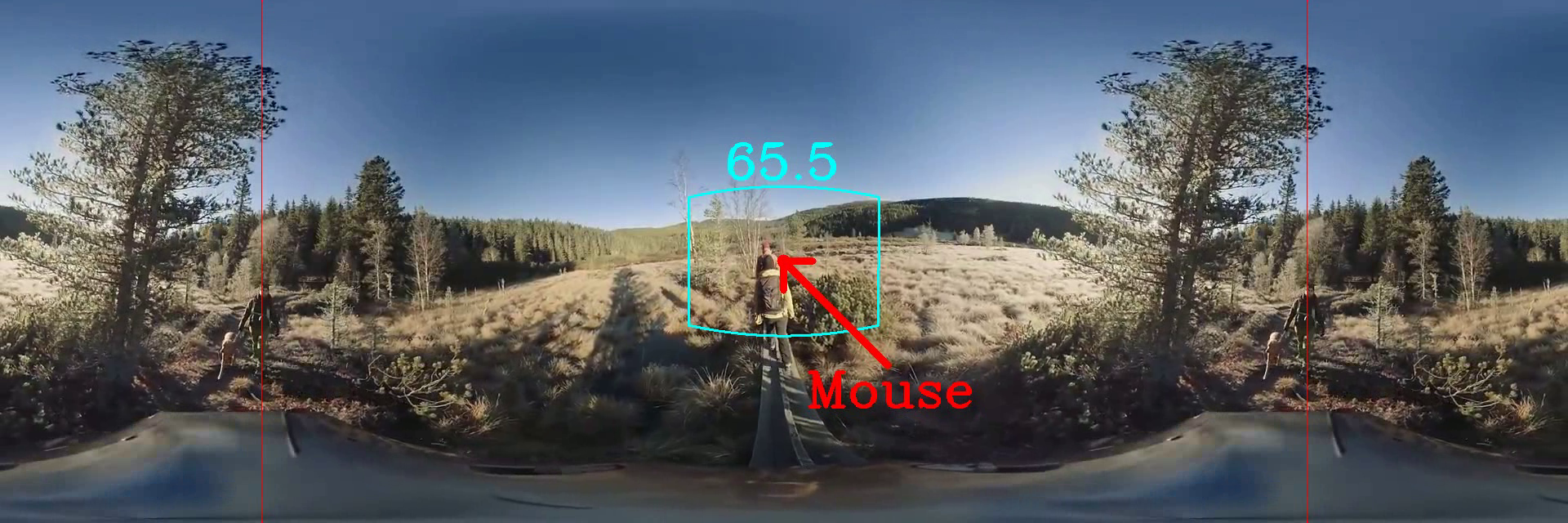}
        \caption{Original.}
        \vspace{8pt}
    \end{subfigure}
    \begin{subfigure}{\linewidth}
        \includegraphics[width=\linewidth]{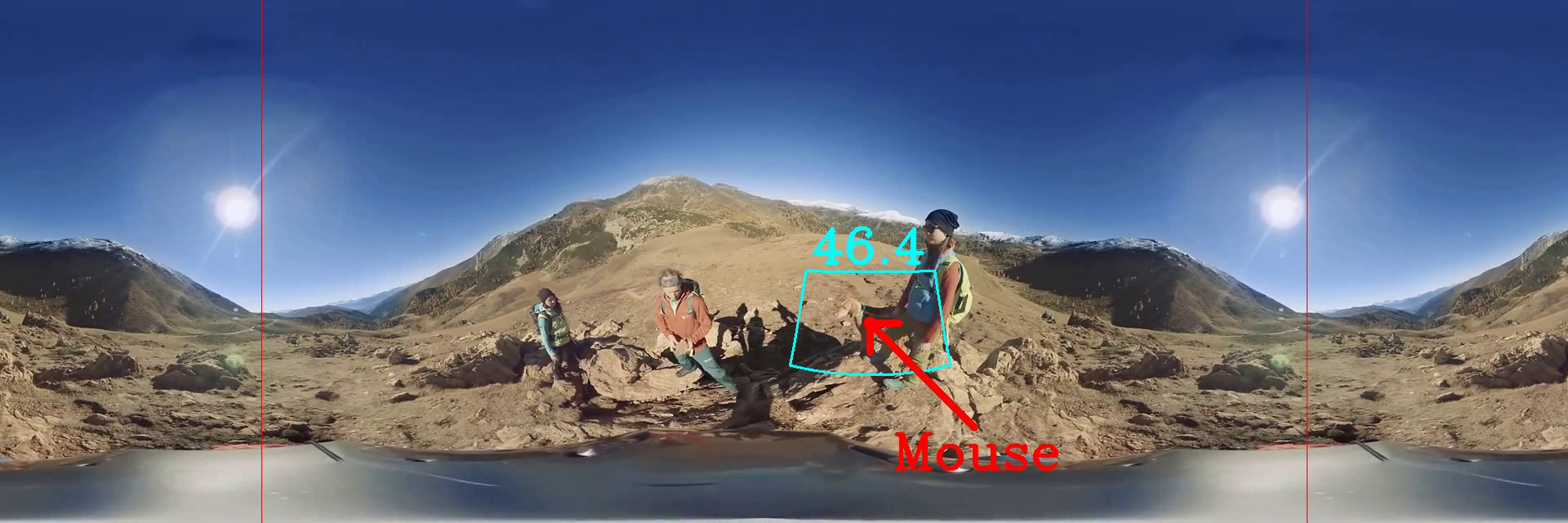}
        \caption{Zoom in.}
    \end{subfigure}
    \caption{
        The interface allows the human editors to control the FOV.
    }
    \label{fig:interface_zoom}
\end{figure*}

The annotation interface displays the $360\degree$ video in equirectangular projection so the editors can see all the visual content at once.
The interface also extends the panoramic strip by $90\degree$ on both sides to mitigate problems due to discontinuities at the edge.
See Fig.~\ref{fig:interface_angle}.
The editors are instructed to move the cursor to direct a virtual NFOV camera,
where the frame boundaries are backprojected onto the panoramic strip in real time to help the editors see the content they capture.
The editors can also control the focal length of the virtual camera.
The available focal lengths are the same as those available to the algorithm,
and the interface will switch to the next available focal length when the editor presses the button for zoom in/out.
See Fig.~\ref{fig:interface_zoom}.

For each $360\degree$ video,
we ask the editors to watch the full video in equirectangular projection first to familiarize themselves with the content.
Next, we ask them to annotate \emph{four} camera trajectories per video.
For each of the four passes,
we pan the panoramic strip by the angle of $[0\degree, 180\degree, 0\degree, 180\degree]$ to force the editors to consider the trajectories from different points of view.
Finally, for the first two trajectories of the first two videos annotated by each editor,
we render and show the output video to the editor right after the annotation to help him understand what the resulting video will look like.

Also see our project webpage for video examples of the interface in action.

